\documentclass[twocolumn,10pt]{asme2ej}
\usepackage{hyperref}   
\hypersetup{
	colorlinks=true,
	linkcolor=blue,
	citecolor=blue,
	urlcolor=blue,
}
\usepackage{epsfig} 
\usepackage{psfrag}
\usepackage{amsmath}
\usepackage{amssymb}
\usepackage{graphicx}
\usepackage{subfigure}
\usepackage{verbatim}
\usepackage[thinlines,thiklines]{easybmat}
\usepackage{latexsym}
\usepackage[dvipsnames]{xcolor}
\usepackage{cite}
\usepackage{stmaryrd}
\usepackage[numbers,sort&compress]{natbib}
\usepackage{multirow}
\usepackage{textcomp}
	\usepackage[ruled]{algorithm2e}

\usepackage{color,soul}
\sethlcolor{white}
	
\definecolor{shadecolor}{rgb}{1,0.9,0.4}
	
\DeclareMathAlphabet{\mathpzc}{OT1}{pzc}{m}{it}

		\renewcommand{\topmargin}{1.0in} 

	\def\atan2{\mathop{\rm atan2}\nolimits}
	\newcommand{\bs}[1]{\ensuremath{{\boldsymbol{#1}}}}
	\def\atan2{\mathrm{atan2}}

			\newcommand\highlightReference[1]{%
				\expandafter\newcommand\csname highlightReference-#1\endcsname{}%
			}
			\let\oldbibitem\bibitem
			\def\bibitem#1 #2\par{%
				\expandafter\ifx\csname highlightReference-#1\endcsname\relax
				\oldbibitem{#1}#2\par
				\else
				\oldbibitem{#1}\highlight{#2}\par
				\fi
			}
			\newcommand\highlight[1]{\textcolor{blue}{#1}}
			
\title{Biomechanical Comparison of Human Walking Locomotion on Solid Ground and Sand}

\author{Chunchu Zhu, Xunjie Chen, Jingang Yi\thanks{Address all correspondence to J. Yi.}\affiliation{Department of Mechanical and Aerospace Engineering\\
Rutgers, The State University of New Jersey\\ Piscataway, NJ 08854, USA\\
Email: \{chunchu.zhu,xunjie.chen,jgyi\}@rutgers.edu}}
			
\begin{document}
\maketitle


\begin{abstract}
{\it Current studies on human locomotion focus mainly on solid ground walking conditions. In this paper, we present a biomechanic comparison of human walking locomotion on solid ground and sand. A novel dataset containing 3-dimensional motion and biomechanical data from 20 able-bodied adults for locomotion on solid ground and sand is collected. We present the data collection methods and report the sensor data along with the kinematic and kinetic profiles of joint biomechanics. A comprehensive analysis of human gait and joint stiffness profiles is presented. The kinematic and kinetic analysis reveals that human walking locomotion on sand shows different ground reaction forces and joint torque profiles, compared with those patterns from walking on solid ground. These gait differences reflect that humans adopt motion control strategies for yielding terrain conditions such as sand. The dataset also provides a source of locomotion data for researchers to study human activity recognition and assistive devices for walking on different terrains.}
\end{abstract}

\section{Introduction}

The biomechanical study of human locomotion has been a fundamental aspect of understanding human physiology and developing assistive devices and wearable robotic systems. However, the majority of gait analysis research and open-source datasets were performed on level, solid ground~\citep{camargo2021comprehensive}. This focus has yielded significant insights into the basic gait biomechanics but has not encapsulated the full spectrum of challenges and variables present during everyday ambulation. When it comes to real-world locomotion, navigating through granular terrains such as sand can be particularly challenging due to their constantly yielding and shifting nature~\citep{kowalsky2021human}. Sand and other yielding surfaces are prevalent in everyday life, including beaches, playgrounds, and off-road trails~\citep{svenningsen2019effect}. The distinct properties of sand, such as its constantly shifting and deformable nature, present unique challenges for maintaining stability and efficiency during locomotion~\citep{zamparo1992energy}. 

Several research works have provided a comprehensive understanding of how different terrain conditions affect human walking locomotion. The metabolic cost of walking on compliant substrates was investigated in~\cite{grant2022does}, and the study provided a foundation for understanding the energy demands of walking on sand. The work of~\cite{zamparo1992energy} specifically examined the energy cost of walking and running on sand and highlighted the increased physical effort required on yielding surfaces. The mechanical work and energetic cost during walking and running on sand and on a hard surface were compared in~\cite{lejeune1998mechanics}. In~\cite{panebianco2021quantitative}, the authors quantitatively compared the speed, temporal segmentation, and variability of walking on solid ground and sand. These studies underline the importance of terrain in influencing gait mechanics and energy efficiency.

The research work on human gait in natural settings~\citep{kowalsky2021human} and on diverse surfaces~\citep{holowka2022forest} provided a comprehensive picture of the interaction between terrain type, gait parameters, and energy expenditure. Specific adaptations to walking on uneven terrains such as unanticipated steps were observed and corresponding lower limb biomechanical response was analyzed in~\cite{panizzolo2017lower}. Another comprehensive overview of the terrain impacts on human sprint locomotion demonstrated the effects of different surfaces such as natural and artificial turf, and sand. The study in~\cite{xu2015influence} discussed the influence of deformation height on estimating the center of pressure (COP) during walking on level ground and sand, showing that the magnitude of COP changes in the anterior-posterior (AP) and medial-lateral (ML) directions differed in both level and cross-slope conditions.

\begin{figure*}[h!]
	\centering
	\includegraphics[width = 6in]{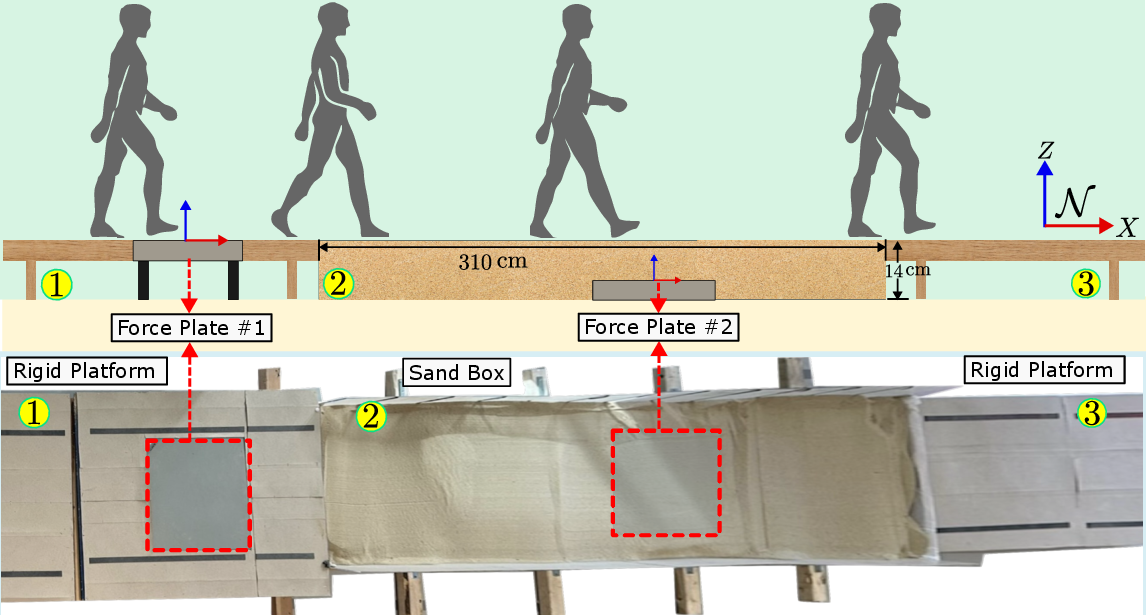}
	\caption{Experimental platform design and setup inside a laboratory for human walking on solid and sand terrains.}
	\label{fig:exp_setup}
\end{figure*}

Recent advancements in assistive technologies have significantly contributed to the understanding of human-terrain interaction during walking locomotion. Developments in exoskeleton control for uneven terrains in~\cite{jatsun2018walking} and the integration of human-in-the-loop control in soft exosuits on different terrains in~\cite{li2022human} have enhanced activity recognition and response adaptability. The use of inertial measurement unit (IMU)-based algorithms for foot-ground contact detection~\citep{kim2017development,TrkovTASE2019} and terrain topography detection~\citep{knuth2023imu} illustrates the complexity of walking dynamics on varied surfaces. These technique developments play a crucial role in real-time walking gait phase estimation, especially on uneven terrain~\citep{medrano2023real}. Additionally, studies focusing on the asymmetry of ground contact during running on sand~\citep{dewolf2019off} and the analysis of ground reaction forces and muscle activity in walking across different surfaces~\citep{jafarnezhadgero2019ground,jafarnezhadgero2022effects} have extended the understanding of the biomechanical adaptations required for locomotion on diverse terrains. Moreover,~\cite{grant2023human} provided a holistic view of how human gait and energetics are altered when traversing substrates of varying compliance.

Understanding the biomechanics of walking on sand has potential implications for various applications, such as rehabilitation, biped/quadruped robot control, and human exoskeleton assistance. The knowledge gained from studying human locomotion on sand can be applied to improve the control and stability of bipedal robots, as well as to enhance the performance of exoskeletons in assisting human movement on diverse surfaces~\cite{ZhuMECC2024}. Therefore, evaluating the biomechanics of walking on sand is essential for advancing our understanding of human locomotion in diverse terrains and developing assistive devices and robotic systems that can adapt to these common environments. All the above-mentioned research emphasizes the complex interplay between human biomechanics and varying terrain types. Comprehensive joint-level biomechanics analysis on locomotion over such yielding terrain however remains scant. Recognizing this deficit, this study introduces a detailed analysis of human locomotion on two distinct terrains: solid ground and sand. The methodology involves collecting and analyzing a novel dataset of biomechanical data from $20$ able-bodied adults. A motion capture system coupled with ground reaction force (GRF) measurements is used to capture both kinematic and kinetic aspects of walking locomotion. We comprehensively analyze joint-level biomechanics, including GRFs and joint torque profiles, in environments that replicate natural walking conditions on both solid and yielding sand surfaces.

The main contributions of this work are twofold. First, an in-depth analysis of the biomechanical characteristics of human walking on granular terrains is presented. To the authors' best knowledge, there is no reported result on GRFs and joint torque profiles and corresponding comparisons for humans walking both on sand and solid ground. This study fills such a knowledge gap and offers new insights into the biomechanical strategies employed by individuals when navigating through granular terrains. The biomechanical analysis in this work not only contributes to enhancing the understanding of human gait mechanics but also serves as a foundation for the development of assistive devices. Second, this work provides an experimental dataset that integrates synchronized motion capture and GRF data. This dataset is expected to be a valuable resource for further biomechanics and wearable sensor data research across varied environmental settings.

\section{Methods}
\label{method}

\subsection{Experimental Setup and Protocol}

In this study, twenty able-bodied healthy participants ($14$ males and $6$ females, age: $24.8\pm 4.0$ years old, height: $171.7\pm 9.5$~cm, weight: $74.5\pm 16.1$~kg) participated in experiments and are labeled as S1 to S20. All participants are self-reported to be in a good health condition. An informed consent form was signed by each participant and the experimental protocol was approved by the Institutional Review Board (IRB) at Rutgers University.

Figure~\ref{fig:exp_setup} shows the experimental setup in a laboratory. A three-segment walkway ($7.5$~m long, $0.76$~m wide, and $0.14$~m high) was constructed as shown in the figure. Segment $1$ was built with reinforced plywood with one embedded force plate (model ACG-O from Advanced Mechanical Technology, Inc., Watertown, MA) to replicate a solid ground condition. Segment $2$ was a sandbox filled with sand to represent a common granular terrain condition. A force plate (from Bertec Corporation) was buried $14$ cm beneath the sand surface to capture the ground reaction forces (GRFs). Although the work in~\cite{xu2015influence} studied the influence of deformation height on the COP, the dissipation of GRFs was not considered. To account for this, the GRF measurements collected within Segment $2$ were calibrated using a correlation based on the sand layer thickness. The calibration process and resulting correlation curves can be found in Appendix~\ref{appendix:calibration}. Segment $3$, made of the same material as segment one, was connected to segment two to ensure participants could comfortably leave segment two without changing their gaits.

The experimental setup is designed to study gait adaptations and changes when traversing different terrains. While this setup differs from previous studies that typically analyze gait on a single, uniform surface for several consecutive strides, this design captures the immediate adaptations and biomechanical adjustments that occur during real-world terrain transitions. To ensure meaningful results, we ignored the first few strides on each terrain and focused our analysis on a single, representative step from each trial, allowing participants to adjust to the new surface conditions before recording the relevant data.

Starting at Segment $1$ of the walkway, each participant was instructed to walk in the preferred manner. For the first $5$ trials, the participant was required to walk back and forth on the walkway to get familiar with walking with the sensing devices and the terrain condition of the walkway. After that, another $6$-$8$ trials of one-directional walking experiments were conducted. The sand surface was flattened before each trial. 

\begin{figure}[h!]
	\hspace{-2mm}
	\includegraphics[width = 3.4in]{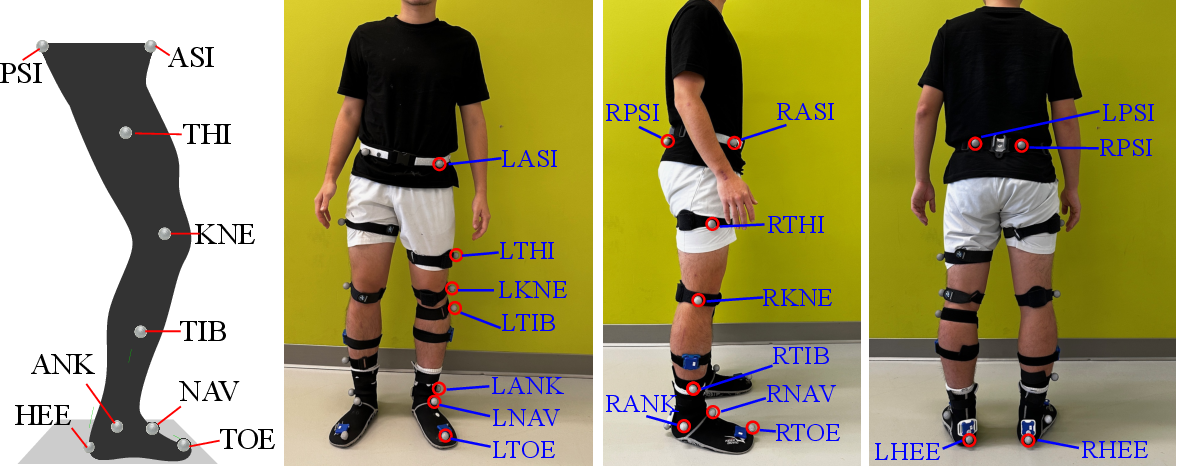}
	\caption{Schematic of the Vicon marker placement for the human lower limb segments. The markers were placed bilaterally.}
	\label{fig:sensors}
	\vspace{-3mm}
\end{figure}

For the walking motion, a motion capture system ($10$ Vantage cameras, Vicon Motion Systems Ltd.) was used to collect lower limb motion information with the sampling frequency $100$~Hz. A total of 18 reflective markers were firmly attached to each participant using straps. Figure~\ref{fig:sensors} shows the specific schematic of Vicon marker positions and their corresponding labels. ``L'' and ``R'' denote ``left'' and ``right'', respectively. Using the marker data, a customized lower-limb model was constructed within the Vicon Nexus software for subsequent analysis. The participants were required to wear a pair of swimming boots for the free movement of foot navicular joints. Meanwhile, the GRF measurements were synchronized via hardware and software setup with the motion data at $1000$ Hz.

\begin{figure*}[h!]
	\centering
	\includegraphics[width = 6in]{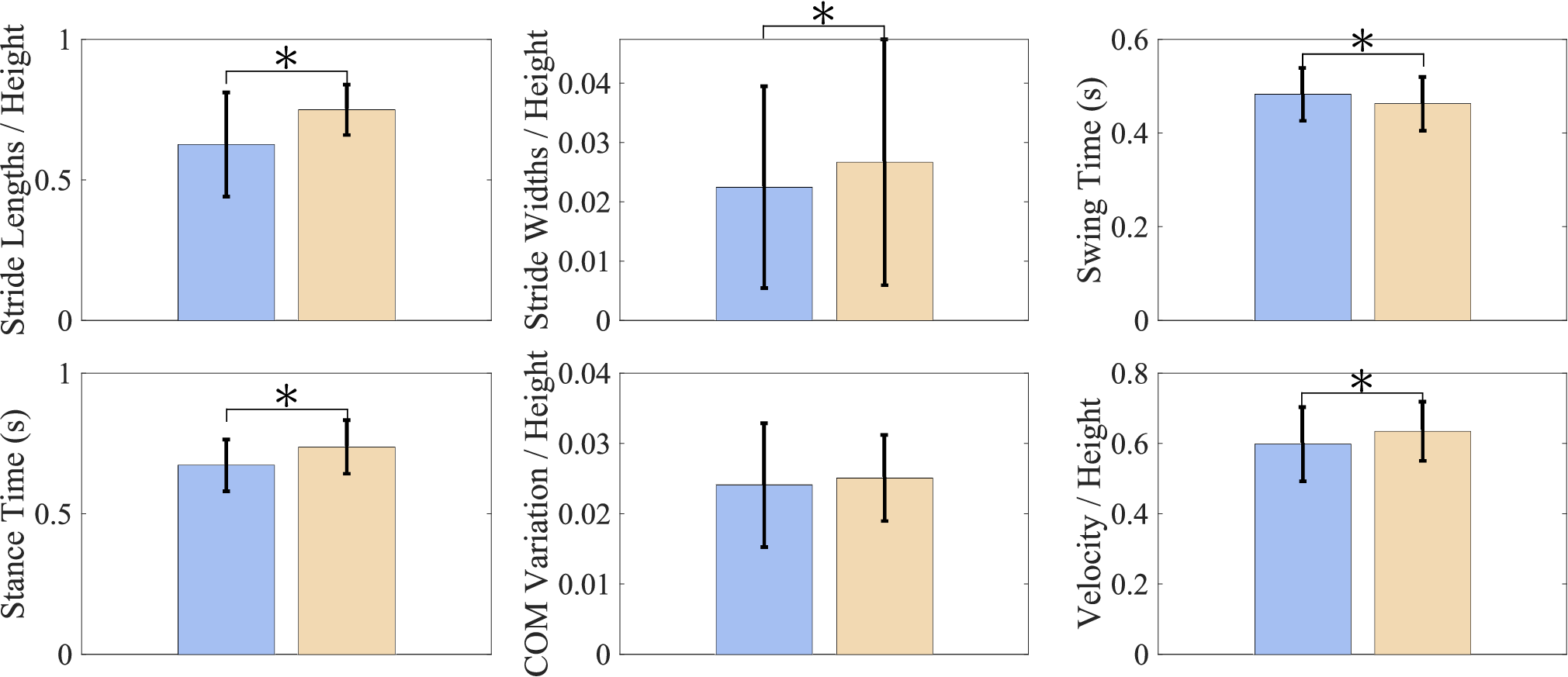}
	\caption{The distribution of step parameters for all participants combined (n = $20$) while walking on the two different terrains: solid ground (blue) and sand (yellow). The comparison was among the stride length, stride width, COM variation, walking velocity (all normalized with respect to participants' heights), swing, and stance time. Data included all strides for individual trials. The label $*$ indicates a significant ($p < 0.05$) difference among trials.}
	\label{fig:stride_comparison}
\end{figure*}

\begin{figure*}[h!]
	\centering
	\includegraphics[width= 6in]{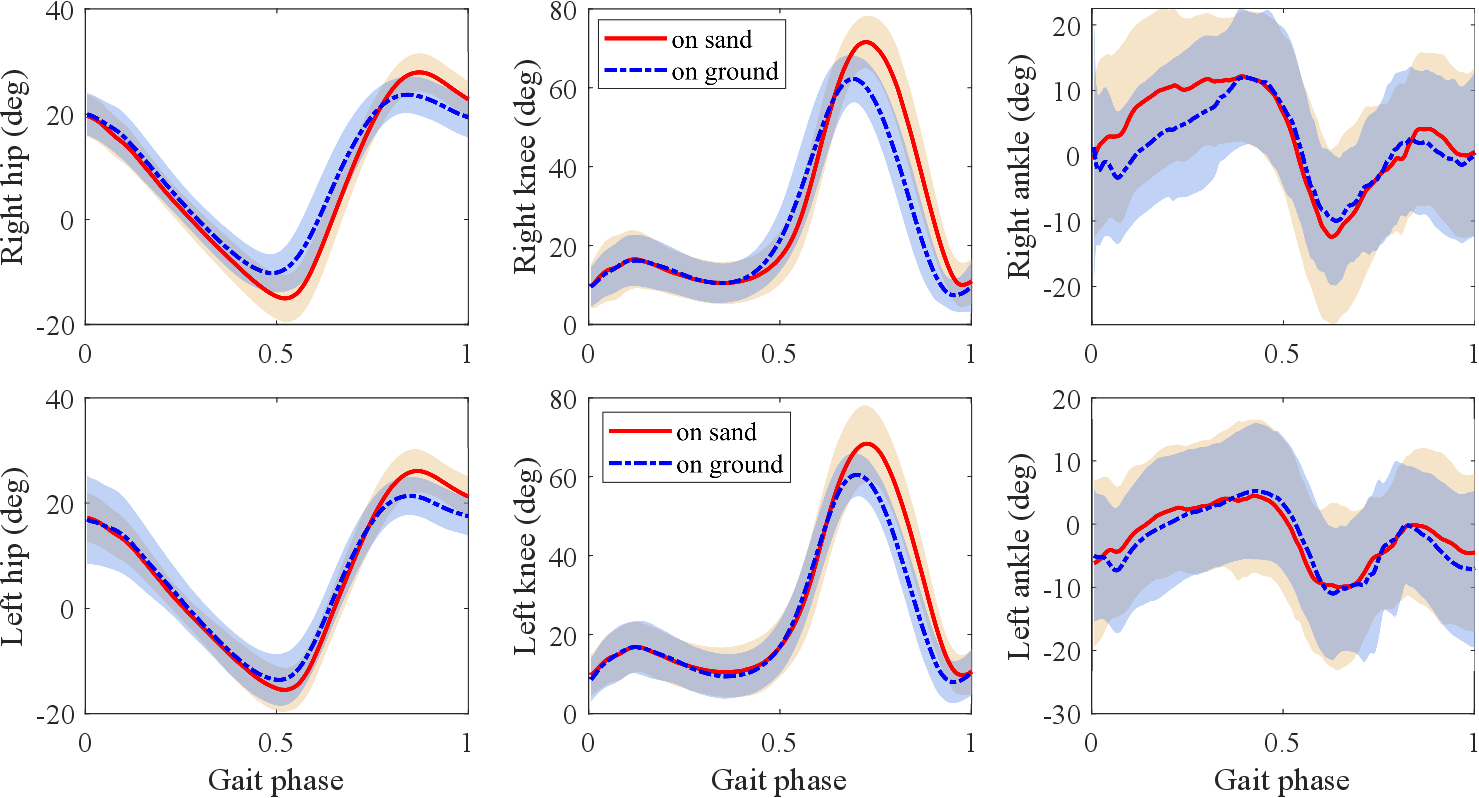}
	\caption{Human walking gait comparison. The top and bottom rows represent the angles of the right and left leg, respectively. The first column: hip flexion(+)/extension(-); the second column: knee flx/extension; the third column: ankle dorsi(+)/plantarflexion(-).}
	\label{fig:Gait_jointAngle}
\end{figure*}

\begin{figure}[h!]
\centering
	\subfigure[]{
		\label{fig:GRF_Fx}
	  \includegraphics[width=3in]{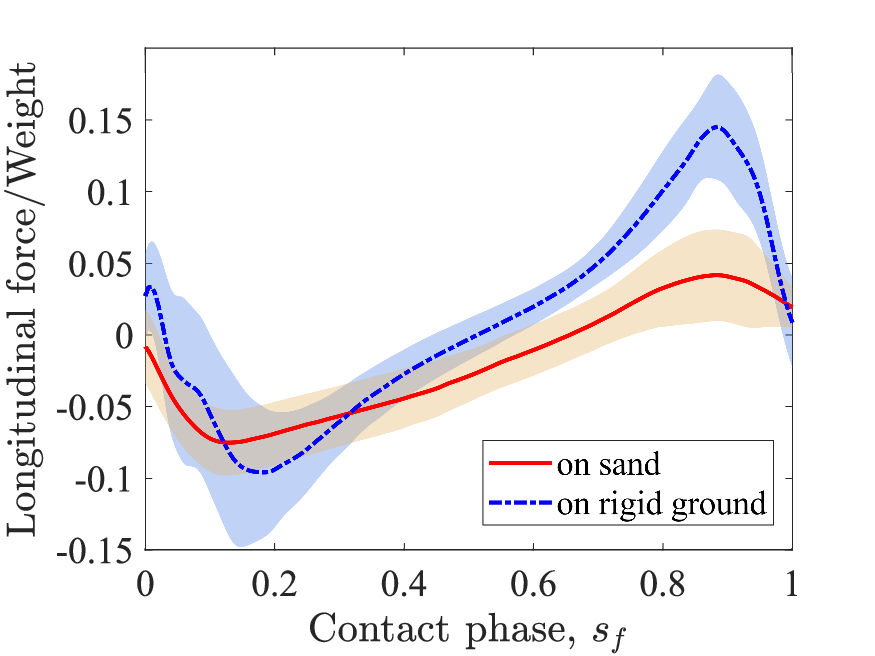}}
\subfigure[]{
	\label{fig:GRF_Fz}
		\includegraphics[width=3in]{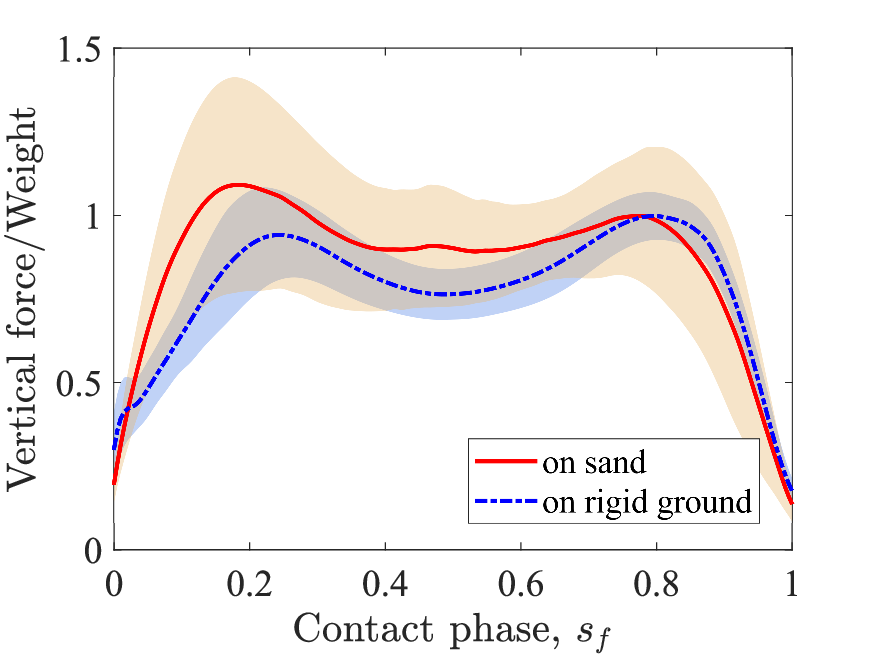}}
	\caption{The comparison of the ground reaction forces on sand and solid ground. (a) Longitudinal forces $F_x$. (b) vertical/normal forces $F_z$.}
\end{figure}

\subsection{Data Processing}

Major gait events such as heel strike and toe-off were defined through motion data to identify the stance phases (i.e., single and double stance phases, respectively) of the gait cycle. The GRFs in the horizontal and vertical directions were smoothed by a moving average filter. The walking gait progression, namely, the joint angle, was normalized by a defined phase variable (between $0$ and $1$) according to the strike events. For clear presentation, the GRFs were normalized the weight of the participant and presented as the functions of the stance variable which was normalized between the heel-strike and toe-off.

The measurements and calculations of interest are the angles of the hip, knee, and ankle joints in the sagittal plane of the human walking locomotion, namely, the $ZOX$ plane shown in Fig.~\ref{fig:exp_setup}. Likewise, we only presented the longitudinal and vertical forces in the $X$-and $Z$-direction, respectively. Joint moments in the sagittal plane were also computed through an inverse dynamics model given the GRFs applied on the foot;  see the detailed description of the joint moments calculation in Appendix~\ref{appendix:ID}. Other kinematic variables such as stride length/width, stance/swing time, vertical variation of center of mass (COM), and average walking velocity highlighted significant variations in the gait patterns between solid ground and sand, and they were included and reported in the study. The data processing was done by using custom algorithms in MATLAB software (Mathworks Inc., Natick, MA). A paired sample t-test was conducted to evaluate the impact of the terrains on the above-mentioned metrics. 

\begin{figure*}[ht!]
	\centering
	\includegraphics[width= 6in]{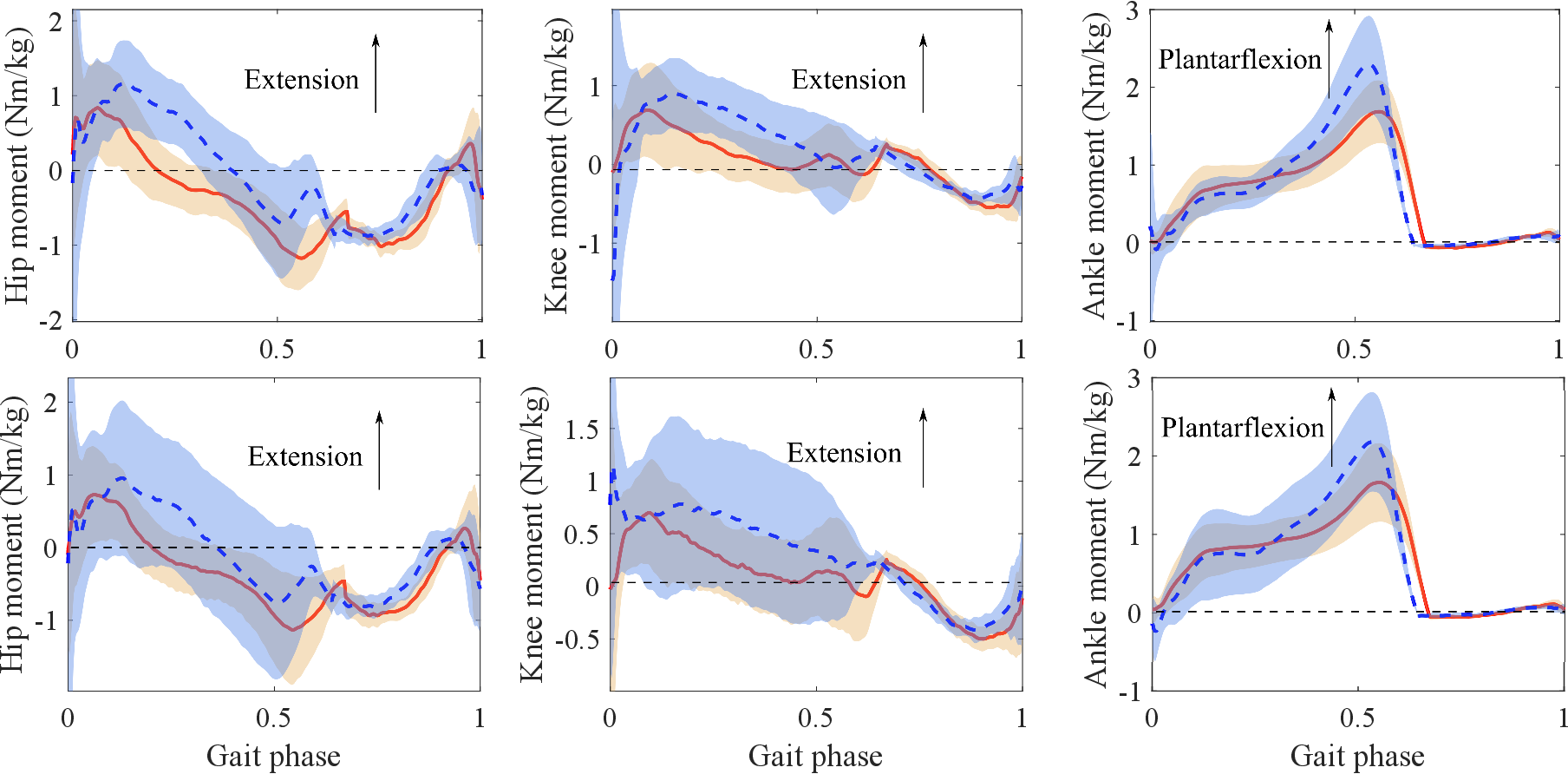}
	\caption{The comparison of the normalized joint moments in the sagittal plane of the human walking locomotion. The top and bottom rows represent the right and left leg, respectively. The extension of the hip and knee moment and plantarflexion of the ankle are indicated as the arrow in the figure.}
	\label{fig:Gait_jointMoment}
\end{figure*}

\section{Experimental Results}
\label{results}

\subsection{Kinematic Analysis Results}

Figure~\ref{fig:stride_comparison} illustrates stride profiles among all the participants. Compared with that on solid ground, stride length was found to be significantly longer (i.e., $20$\%) on sand, with an accompanying increase of $17$\% in stride width. Statistical analysis revealed that walking on sand resulted in a significantly larger normalized stride length ($0.74 \pm 0.08$) compared with on solid ground ($0.663 \pm 0.19$) ($p < 0.05$; $d = 0.53$). Similar changes in normalized stride widths were noticed when walking on sand ($0.026 \pm 0.021$) compared with stable ground ($0.023 \pm 0.017$) ($p < 0.05$; $d = 0.15$). This gait adaptation is likely due to the deformable nature of the sandy terrain. Walking on sand showed a significant increase in both swing time and stance time compared to walking on solid ground. The swing time for walking on sand ($0.46 \pm 0.057$ s) was significantly shorter than that on solid ground ($0.48 \pm 0.056$ s) ($p < 0.05$; $d = 0.26$). Similarly, the stance time for walking on sand ($0.74 \pm 0.09$ s) was significantly longer than that on solid ground ($0.67 \pm 0.09$ s) ($p < 0.05$; $d = 0.57$). Despite these increases in swing and stance times, participants tended to walk faster on sand than on solid ground. The ratio between normalized average walking speed on solid ground ($0.60 \pm 0.11$) versus that on sand ($0.64 \pm 0.08$) was about $0.94$ ($p < 0.05$; $d = 0.54$). The mean COM vertical variation during each step on the sand was also observed to increase by $6$\%, although this difference was not statistically significant ($p > 0.05$; $d = 0.01$). This is possibly due to the extra effort from the joints when participants walked on the sand. The gait phase percentages reveal slight differences between walking on sand and solid ground, with the stance phase taking up $42\%$ of the gait cycle on solid ground and $38\%$ on sand.

Differences in joint angles, particularly in the lower extremities, were observed, indicating a distinct adjustment in walking strategy on sand compared to solid ground. Figure~\ref{fig:Gait_jointAngle} shows the profile comparison for the hip, knee, and ankle joint angles throughout the gait cycle. Although there were no clear differences in the overall joint angle patterns, statistical analysis revealed significant differences in peak joint angles. The peak hip flexion angle was significantly greater on sand ($29.36 \pm 3.84^\circ$) than on solid ground ($p < 0.05$; $d = 0.11$). Similarly, the peak knee flexion angle was significantly greater on sand ($75.21 \pm 7.96^\circ$) compared to solid ground ($p < 0.05$; $d = 0.37$). The peak ankle dorsiflexion angle was also significantly greater on sand ($18.29 \pm 7.50^\circ$) than on solid ground ($p < 0.05$; $d = 0.16$). These findings highlight the specific adaptations in lower limb joint kinematics when walking on sand compared to solid ground.

\subsection{Kinetic Analysis Results}

Figures~\ref{fig:GRF_Fx} and~\ref{fig:GRF_Fz} show the longitudinal and vertical GRFs normalized by the participant's weight, respectively. The shaded region indicated the one-standard variation. It is observed that the normalized magnitude of the maximum longitudinal force $F_x$ on sand is relatively smaller than that on solid ground in both forward ($0.0464 \pm 0.03$ vs. $0.15 \pm 0.035$) and backward directions ($-0.076 \pm 0.024$ vs. $-0.1 \pm 0.047$). However, regarding the vertical force $F_z$, sandy terrain might provide more supporting force during the heel strike ($1.22 \pm 0.22$ vs. $1.03 \pm 0.084$) compared with the solid ground. It is also noticeable that there is a slight difference with the double-hump pattern of the vertical force $F_z$ throughout the contact phase. The magnitudes of two humps on the solid ground appear almost equivalent. Nevertheless, for the sandy terrain, the first hump of the force is slightly higher than the second one in the late contact phase. Similar results were also reported on the level sand in~\cite{xu2015influence}. The significant differences in longitudinal ($p < 0.05$; $d = 0.58$) and vertical ($p < 0.05$; $d = 0.36$) GRFs between sand and solid ground indicate that the type of terrain has a substantial impact on the forces exerted by the foot during walking. The increased variability and different magnitudes of GRFs on sand compared to solid ground suggest that walking on sand requires different biomechanical strategies, possibly due to the less stiff surface.

\begin{figure}[h!]
	\centering
	\includegraphics[width=3.3in]{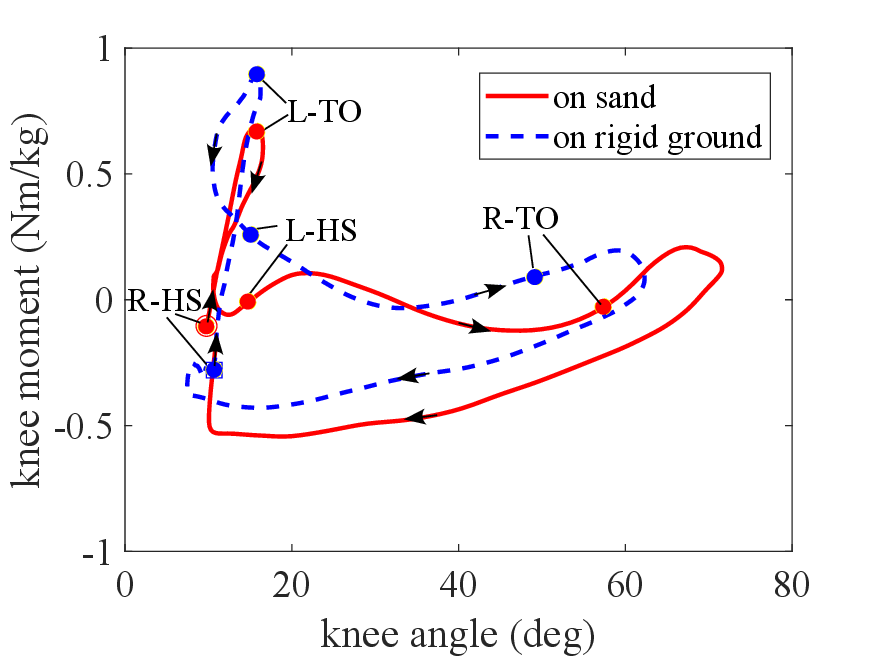}
	\caption{The comparison of the knee biomechanics over a strike cycle. Heel-strike (HS) and toe-off (TO) events are indicated. The knee joint angles and moments on sand and solid ground are all represented by the average values. }
	\label{fig:Knee_jointMoment}
\end{figure}

Figure~\ref{fig:Gait_jointMoment} illustrates the joint moments in the sagittal plane normalized by the participant's mass. The inverse dynamics model was discussed in~\cite{shamaei2013estimation} and the detailed calculation is presented in Appendix~\ref{appendix:ID}. From the comparison results, we observe that for able-bodied locomotion, the joint moments kept consistent bilaterally both on sandy and solid terrains. Figure~\ref{fig:Knee_jointMoment} shows the comparison of the knee stiffness profiles. Compared with that on solid ground, the knee moment contour of walking on sand shifts towards the large angle amplitude and lower moment amplitude directions. However, for the characteristic feature such as knee stiffness defined in the stance phase~\citep{shamaei2013estimation}, i.e., shown as the trajectory R-HS$\rightarrow$L-TO$\rightarrow$L-HS in the figure, did not change significantly. The average knee flexion stiffness values ($K_f$) during the R-HS$\rightarrow$L-TO phase are approximately 0.15 Nm/(deg$\cdot$kg), while the average knee extension stiffness values ($K_e$) during the L-TO$\rightarrow$L-HS phase are around 0.17 Nm/(deg$\cdot$kg) on both solid ground and sand. However, for the swing phase (R-TO$\rightarrow$R-HS), the knee stiffness values shift due to the difference in knee angles, although the average magnitudes remain similar at about 0.011 Nm/(deg$\cdot$kg) on both terrains.

\section{Discussion}
\label{dis}

In this study, we examined human walking locomotion across two different terrain conditions, with an emphasis on a granular sandy condition. We collected walking data and extracted detailed kinematic and kinetic measurements and calculations. Moreover, we encompassed a set of comparisons throughout the walking gait profile (i.e., joint angles), joint moments, and GRFs. Insights on walking locomotion on sandy terrain are discussed as follows.

\subsection{Kinematics Features}

Different from the experiment setup designed in~\cite{svenningsen2019effect}, the participants in this study were asked to walk across two different terrains sequentially from the solid ground to the sand box, allowing each participant to keep the preferred comfortable walking pace. Our findings suggest that walking strategies adapt significantly when transitioning from solid ground to sand. The extended stride length and width observed on the sand were likely compensatory mechanisms to maintain balance on the yielding surface. Participants demonstrated a higher average walking speed on sand compared to solid ground; see the velocity ratio result in Fig.~\ref{fig:stride_comparison}. This result is consistent with~\cite{panebianco2021quantitative}, which reported that the self-selected speed on a $60$~m distance increased when walking on sand with respect to solid ground, with the highest speed on wet sand. This observation suggests that individuals might compensate for the challenging terrain by adopting a faster pace, potentially to minimize the time spent on an unstable surface. Additionally, the increased COM vertical variation indicates an adaptive response to optimize energy efficiency and maintain stability on the yielding terrain, which was also found for the mid-stance phase ($21$-$60$\% gait cycle) in~\cite{svenningsen2019effect}. These adaptations are pivotal for navigating through granular substrates, where it is more challenging than navigating on solid ground.

The joint angles, especially in the lower extremities, demonstrate distinct gait variations on solid and sand terrains. Walking on the sandy terrain results in greater dorsiflexion of the ankle in the early stance phase but a similar peak amplitude of the ankle plantar flexion near toe-off. This is because the heel would experience an intrusion into the sand and then the forefoot was prevented from plantarflexion at the beginning of the stance phase; see the small portion of ankle plantarflexion on the solid ground shown as the blue dashed line in Fig.~\ref{fig:Gait_jointAngle}. Furthermore, a relatively larger variability of the ankle dorsi/plantarflexion on the sand was found compared with the hip flexion/extension and knee flex/extension. This represents the nature of complexity and instability of deformable terrain locomotion. 

The greater flexion observed in the hip and knee joints during the toe-off and swing phases on sand can be interpreted as a compensatory mechanism for the loss of momentum during the stance phase. The deformable nature of sand reduces the efficiency of energy transfer, necessitating additional joint flexion to facilitate foot clearance and forward progression. These kinematic findings underscore the adaptability and fine-tuned control required for efficient and stable locomotion on deformable surfaces. Understanding these adaptations can inform the design of assistive devices, prosthetics, and rehabilitation protocols, as well as guide the development of robotic systems and control algorithms that can effectively traverse challenging terrains.

\subsection{Kinetics Characteristics} 

The kinetic analysis revealed critical insights into the biomechanical demands of walking on different terrains. For kinematic profiles such as joint moments, compared with on solid ground, sand walking locomotion required less hip extensor action until the mid-stance to control knee flexion and the forward rotation of the upper body. The longitudinal force shown in Fig.~\ref{fig:GRF_Fx} has a small backward component at the hip joint, which requires small hip extensors consequently. However, compared to solid ground, walking on sand demanded greater hip flexor to pull the thigh forward and upward at the latter half of the stance phase (including the double stance) and to get ready for the knee extension and swing. This indicates large energy consumption to compensate for momentum loss at the stance leg on sand. The distinct patterns in GRFs and joint moments suggest that walking on sand necessitate a reorientation of muscular efforts and joint mechanics. This reorientation is a complex interplay between maintaining stability, managing energy efficiency, and adapting to the terrain's physical characteristics. 

The differences in GRFs between sand and solid ground highlight the significant impact of terrain type on the forces exerted by the foot during walking. The relatively small normalized magnitude of the maximum longitudinal force on sand suggests that the yielding nature of sand reduces the ability to generate propulsive and braking forces, possibly due to the dissipation of energy as the foot interacts with the deformable surface. In contrast, the vertical force on sandy terrain exhibits a higher peak during the heel strike phase and a different double-hump pattern compared to solid ground, indicating that the deformable surface significantly alters the loading pattern during the stance phase. The increased variability and different magnitudes of GRFs on sand also suggest that walking on this surface requires adapted biomechanical strategies to maintain stability and forward progression. These adaptations could involve changes in muscle activation patterns, joint stiffness, and overall gait mechanics to ensure efficient and safe locomotion on the challenging surface. 

The knee biomechanical profile potentially provides a tool for assistive device controllers to adjust strategies accordingly. By leveraging joint torque and angle data, assistive controllers can provide phase-specific assistance that aligns with the natural gait cycle, enhancing movement synergy. For example, using the knee exoskeleton in~\cite{ZhuRAL2023}, we can further incorporate the active knee biomechanics profile with respect to knee flexion/extension. Adaptive real-time biomechanical feedback control can dynamically adjust assistance on varying terrains and, therefore, improve stability and energy efficiency for the user. This data-driven approach promises to refine the integration of exoskeletons with human locomotion and potentially enhance assistive device functionality~\cite{ZhuMECC2024}.

This study represents an advanced step in analyzing biomechanics during human locomotion across varying terrain types, specifically solid ground and sand. It is the first study that reports the GRFs and joint moments for humans walking on yielding terrains such as sand. The locomotion dataset provided by this study also generates new knowledge and enables to use wearable assistive devices and IMU-based activity recognition for human locomotion on yielding terrains. The kinematic analysis of the lower limbs aligns with existing studies, such as~\cite{van2017effect}. In contrast to~\cite{xu2015influence}, which posited an experimental design that minimizes force dissipation in the sand, the calibration results detailed in Appendix~\ref{appendix:calibration} indicate a discrepancy between forces at the sand surface and the GRFs measured by the embedded force plate. Despite requiring calibration for magnitude, the reported longitudinal and vertical GRFs in this work are in agreement in pattern to those reported by~\cite{xu2015influence}. Contrary to~\cite{svenningsen2019effect}, which found reduced walking speed and stride length on sand, the participants in this study exhibited increased pace, longer strides, and greater vertical COM displacement. This discrepancy is attributed to the consistent starting conditions for all trials, eliminating bias from self-selected walking strategies per terrain. 

The methodology in this work thus captures the actual transition between gaits when moving from solid ground to sand. The adaptive locomotion on the sand can be ascribed to the inherent compensatory mechanisms of the human motor system by adjusting stride and body mechanics to maintain balance and stability on the less stable sand surface. The 7.5-meter walkway used in this study, although relatively short, provides valuable insights into the immediate adaptations and biomechanical changes that occur when transitioning between solid ground and sand. The platform design allows for the capture of at least one complete nature gait cycle on each terrain type, enabling the analysis of important biomechanical parameters. This study design is particularly useful for understanding the strategies employed by the human locomotion system when confronted with sudden changes in surface properties, providing valuable insights into the mechanisms of gait adaptation on varying surfaces.

While acknowledging some study limitations, their impact on our findings is not significant. The study's design permitted participants to walk at their preferred pace, leading to variability in walking speeds, diverging from other studies such as~\cite{camargo2021comprehensive} focused on locomotion velocity effects. This variability might influence biomechanical data consistency. In~\cite{xu2015influence}, the results showed the magnitude of COP changes on level ground in the anterior-posterior (AP) direction is within 4 mm when the sand deformation is 20 mm. Therefore, for simplicity, we take the COP position from the GRF data. Additionally, the research scope was limited to straight walking on level sand and solid ground, excluding complex movements like ascending, descending, and turning on granular terrains. Such limitations might restrict a comprehensive understanding of biomechanical adaptations in gait activities. The dataset lacks other lower limb biomechanics, such as EMG signals, important for analyzing muscle activation patterns. Future research should include diverse locomotive modes and participant demographics, using an array of sensors for gait analysis on different terrains to further enhance our understanding of human locomotion on granular terrains. 

\section{Conclusion}
\label{con}

This study has contributed to the understanding of human walking locomotion on granular terrain by providing a detailed biomechanical analysis of gait adaptations. The findings were derived from comprehensive biomechanical and wearable sensor data from $20$ able-bodied adults and highlighted the intricate adjustments in stride, joint mechanics, and ground reaction forces necessary for efficient movement on challenging surfaces such as sand. These insights offered valuable implications for the development of advanced assistive devices and responsive robotic systems. We also provided and shared open-source kinematic and kinetic dataset and a comparison of human walking on solid and sand surfaces. This research work provided biomechanics dataset and methodology for future studies to explore a broader range of locomotion conditions and participant demographics and therefore, enhance the applicability of wearable sensing and assistive technologies in diverse environmental settings.

\section*{Acknowledgments}

The authors thank Mr. Aditya Anikode of Rutgers University for his help in constructing the experimental setup in this study. The work was partially supported by US National Science Foundation (NSF) under award CMMI-2222880.

\vspace{3mm}

\appendix
\noindent{\bf Appendices}
\vspace{-5mm}

\section{Force Calibration}
\label{appendix:calibration}
  
The force plate and sandbox configurations in the experiment were similar to that in~\cite{xu2015influence}. However, the dissipation of the force under this design cannot be negligible. During the walking experiments, the force plate was embedded beneath a layer of sand with a depth of $14$~cm. The GRFs observed at the sand's surface inherently differed from those recorded by the force plate because of the deformation characteristics and energy dissipation properties of sand substrates. Consequently, it is necessary to conduct force calibration to obtain accurate GRFs from the force plate.

\begin{figure}[h!]
	\vspace{1mm}
	\hspace{-1mm}
	\includegraphics[width = 3.35in]{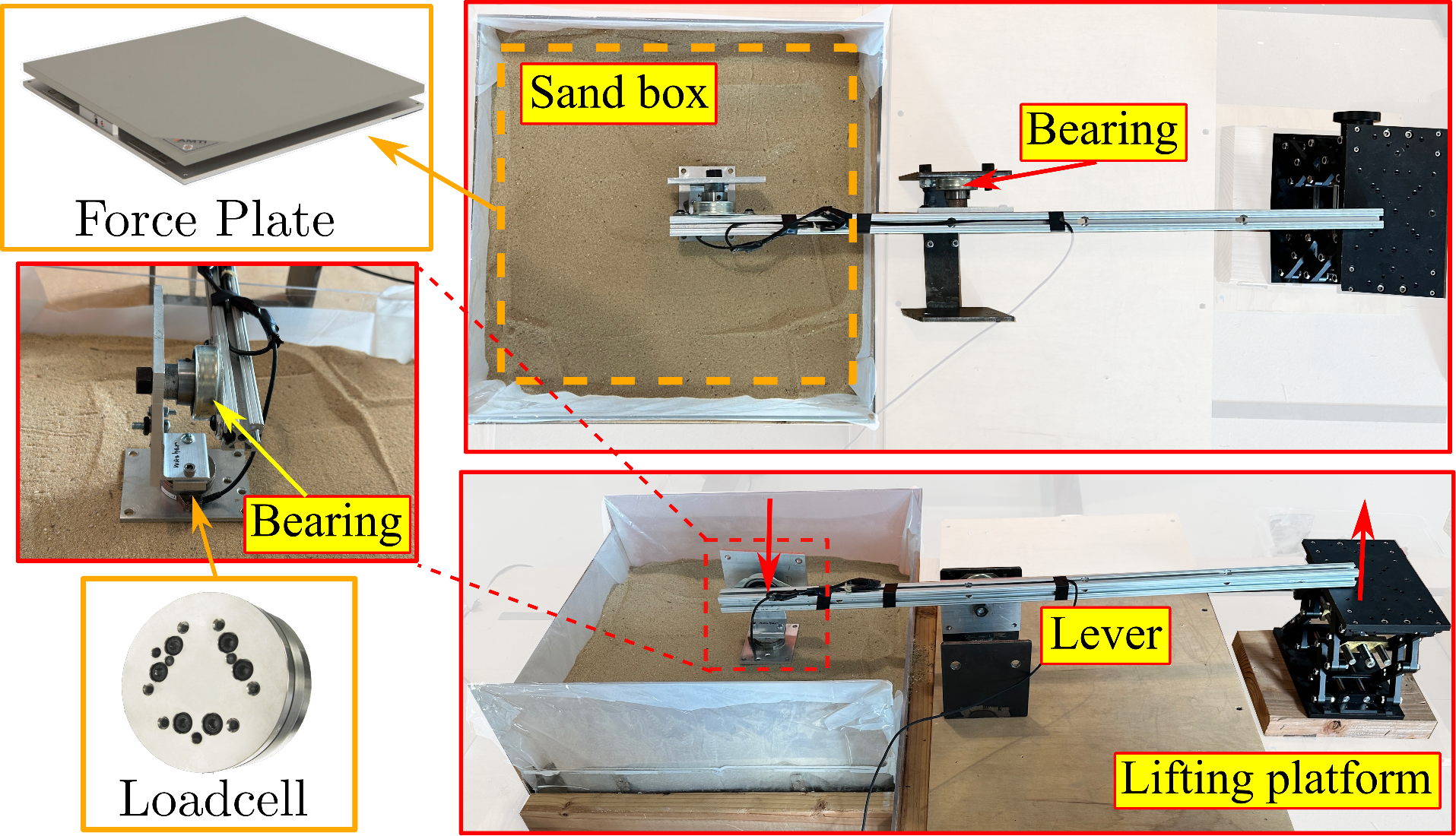}
	\caption{Experimental setup for force calibration on sand terrain.}
	\label{fig:calibrationSetup}
\end{figure}

As shown in Fig.~\ref{fig:calibrationSetup}, we designed and built a sand box with a force plate positioned beneath the sand layer. The calibration setup consists of a lever mechanism, featuring a force/torque load cell (ATI mini45) at one end and a vertically adjustable displacement lift platform at the other end. By raising the platform and leveraging the mechanical advantage of the bearing structure, the compression plate firmly generated a controlled vertical force onto the sand's surface. The magnitude of the applied force was recorded by the load cell and data collection was synchronized with the force plate readings. This setup allowed us to establish a correlation between the force exerted at the surface and the force recorded by the embedded force plate under varying sand thickness conditions.

The calibration was performed by applying static known forces (obtained from the load cell and denoted by $F_{z}^{s}$) and recording the corresponding force plate measurements (denoted by $F_{z}^{b}$) at different sand thicknesses. {The forces were exerted from $0$ to $200$~N with a $25$~N increment, and then lowered to $0$ with the same decrement.} We increased the sand depth from 0 to $14$~cm by the increment of $1$~cm. The sand surface was paved flat before each compression calibration. These measurements were then used to generate a force ratio curve ($\zeta = F_{z}^{b}/F_{z}^{s}$) that served as a calibration reference for interpreting the GRFs during the walking trials. This curve was essential for compensating for the deformation of sands and energy dissipation at various depths and this would allow for a precise adjustment in our biomechanical analysis. Moreover, it was confirmed that the location had no significant impact on the force plate measurements for the area directly above the force plate.

\begin{figure}[h!]
	\hspace{-1mm}
	\includegraphics[width=3.35in]{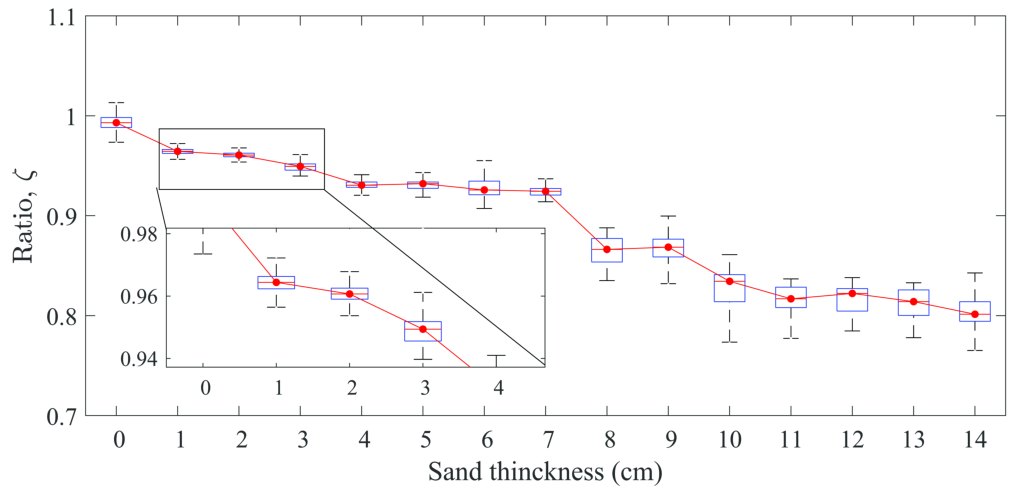}
	\caption{The force calibration ratio results for force plate embedded at different sand depths.}\label{fig:forceCali_results}
\end{figure}

Figure~\ref{fig:forceCali_results} shows the force calibration results. The force measured beneath the sand was indeed smaller than that at the surface. Overall, the force ratio $\zeta$ decreases as the sand layer becomes thicker. A significant ratio drop is observed when the sand thickness rises from $7$ to $8$~cm. In summary, this correction force ratio was used to re-calculate the actual GRFs from the force plate measurements. For the participant walking experiments, we used the ratio $\zeta=0.81$ for the $14$-cm sand depth, and therefore, the real vertical ground reaction force $F_z^{s}=F_z^{b}/\zeta$, where $F_z^{b}$ is the force plate measurements in the vertical direction.

\section{Inverse Dynamics}
\label{appendix:ID}

In this section, we present the calculation of the joint moments in the sagittal plane based on inverse dynamics. The formulations for ankle, knee, and hip joint moments are derived and articulated using Newtonian-Euler formulation specific to the foot, shank, and thigh segments, respectively. Similar derivations can be found in the support materials from the work of~\cite{shamaei2013estimation}.

\begin{figure}[h!]
	\hspace{-1mm}
	\includegraphics[width=3.35in]{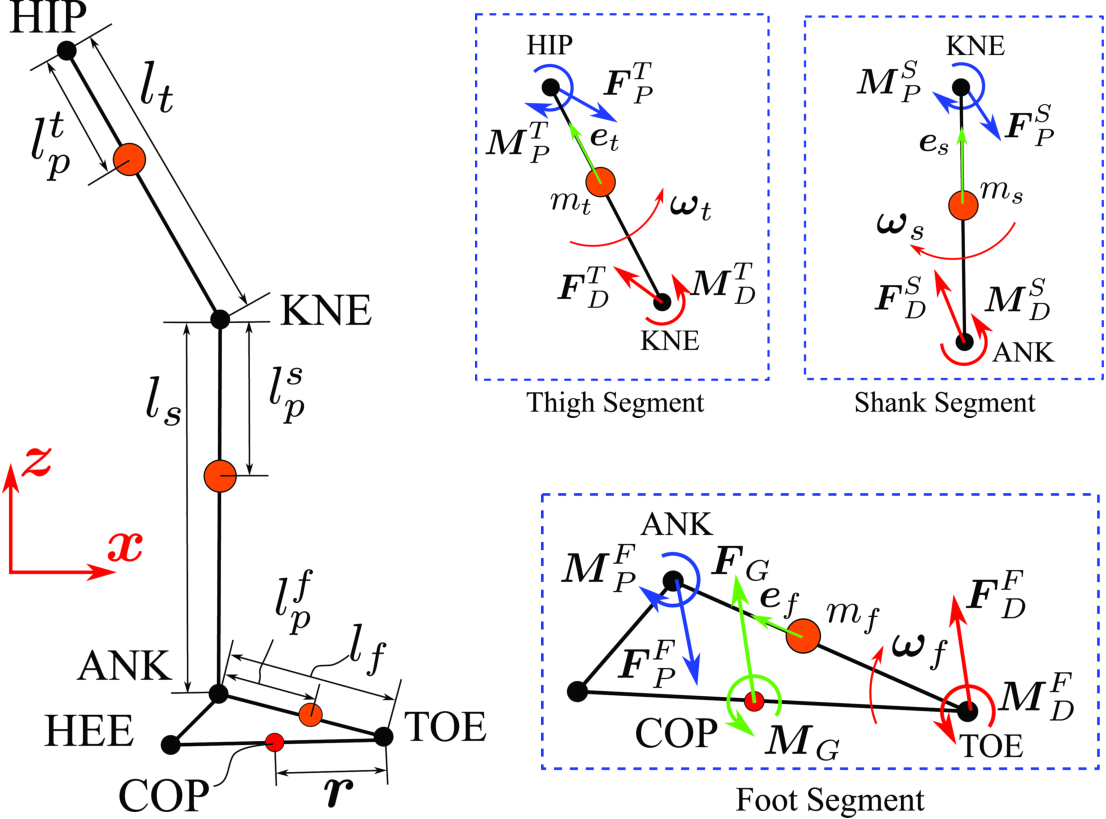}
	\caption{The schematic of the leg for the inverse dynamics in the sagittal plane. Each segment is considered as two joints, namely proximal and distal joints, respectively.}\label{fig:InverseDyns}
\end{figure}

As shown in Fig.~\ref{fig:InverseDyns}, the leg consists of three segments, namely, the thigh, shank, and foot segment. For each segment, there are two joints, proximal and distal joints. To obtain the joint moments of each segment, the Newtonian-Euler method is used. Since the procedure is consistent for each segment, we only present the derivation of the proximal and distal joint moments for the foot segment as a brief example. For the foot segment, the proximal joint is the ankle joint and the distal one is the toe, and the ground reaction force vector $\boldsymbol{F}_G$ and moment vector $\boldsymbol{M}_G$ are applied at the COP point.

First, $\boldsymbol{F}_G$ and $\boldsymbol{M}_G$ should be transferred to the distal joint (toe) such that
\begin{equation}\label{eqn:apndx_toe}
  \boldsymbol{F}_D^F = \boldsymbol{F}_G, ~\boldsymbol{M}_D^F = \boldsymbol{M}_G-\boldsymbol{r} \times \boldsymbol{F}_G,
\end{equation}
where $\boldsymbol{F}_D^F$ and $\boldsymbol{M}_D^F$ are the distal force vector and moment vector at the toe, respectively. $\boldsymbol{r}$ represents the position vector from the COP to the toe.

Next, the forces on the foot segment should satisfy $\Sigma \boldsymbol{F}=m_f \boldsymbol{a}_f$, where $m_f$ is the mass of the foot segment and $\boldsymbol{a}_f$ is the acceleration of the foot. Therefore, the force at the proximal joint (i.e., the ankle) is
\begin{equation}\label{eqn:apndx_ankleForce}
  \boldsymbol{F}_P^F = -\boldsymbol{F}_D^F + m_f \boldsymbol{a}_f - m_f g \boldsymbol{e}_z,
\end{equation}
where $\boldsymbol{F}_P^F$ is the proximal joint force vector and $\boldsymbol{e}_z$ is the unit vector along the $Z$ axis.
Using the Euler equation of the foot segment $\Sigma \boldsymbol{M}_f = I_f \boldsymbol{\dot{\omega}}_f$, where $I_f$ is the mass moment of inertia about the center of mass of the foot segment and $\boldsymbol{\dot{\omega}}_f$ is the angular acceleration of the foot segment. By plugging known forces, we obtain
 \begin{equation}\label{eqn:apndx_ankleMoment}
  \boldsymbol{M}_P^F = -\boldsymbol{M}_G - \left(\boldsymbol{r} + l_f \boldsymbol{e}_f\right) \times \boldsymbol{F}_G - l_p^f \boldsymbol{e}_f \times m_f\left(\boldsymbol{a}_f + g\boldsymbol{e}_z\right) + I_f \boldsymbol{\dot{\omega}}_f,
\end{equation}
where $\boldsymbol{M}_P^F$ is the proximal joint moment, namely, the ankle moment. $l_f$ and $l_p^f$ are the length of the foot segment and distance from the ankle to the center of mass of the foot segment, respectively. $\boldsymbol{e}_f$ is the unit vector of the foot segment direction (shown in Fig.~\ref{fig:InverseDyns}) that represents the orientation of the foot.

For the knee and hip joint moments, we follow the same process as we treat the foot segment. For instance, we take the ankle and knee joint moments as the distal and proximal joint moments, respectively, for the shank segment. The knee and hip joint moments are then treated as the distal and proximal joint moments, respectively, for the thigh segment. Then, the knee and hip moments are calculated respectively as
\begin{align}
\bs{M}_P^S = &-\boldsymbol{M}_G - \left(\boldsymbol{r} + l_f \boldsymbol{e}_f + l_s\boldsymbol{e}_s \right) \times \boldsymbol{F}_G +  I_f \boldsymbol{\dot{\omega}}_f + I_s \boldsymbol{\dot{\omega}}_s- (l_p^f \boldsymbol{e}_f + \nonumber \\
    & l_s\boldsymbol{e}_s) \times m_f\left(\boldsymbol{a}_f + g\boldsymbol{e}_z\right)-l_p^s \boldsymbol{e}_s \times m_s\left(\boldsymbol{a}_s + g\boldsymbol{e}_z\right),
\label{eqn:apndx_kneeMoment}
\end{align}
and 
\begin{align}
    \bs{M}_P^T =&-(\bs{r} + l_f \bs{e}_f + l_s\bs{e}_s + l_t\bs{e}_t) \times \bs{F}_G-(l_p^s \bs{e}_s + l_t\bs{e}_t) \times \nonumber \\ 
	&  m_s(\bs{a}_s +g\bs{e}_z)- \left(l_p^f \boldsymbol{e}_f + l_s\boldsymbol{e}_s +l_t \boldsymbol{e}_t\right) \times m_f\left(\boldsymbol{a}_f + g\boldsymbol{e}_z\right) \nonumber \\
	& + I_f \boldsymbol{\dot{\omega}}_f + I_s \boldsymbol{\dot{\omega}}_s + I_t \boldsymbol{\dot{\omega}}_t -\boldsymbol{M}_G ,
\label{eqn:apndx_hipMoment}
\end{align}
where $\boldsymbol{e}_s$ ($\boldsymbol{e}_t$) is the unit vectors of the shank (thigh) segment. $l_s$ ($l_t$) is the length of the shank (thigh) segment. $l_p^s$ ($l_p^t$) is the distance from the knee (hip) to the center of mass of the shank (thigh) segment. $I_s$ ($I_t$) is the moment inertial about the center of mass of the shank (thigh) segment. The angular acceleration of the shank (thigh) segment is denoted as $\boldsymbol{\dot{\omega}}_s$ ($\boldsymbol{\dot{\omega}}_t$).

The formulations \eqref{eqn:apndx_ankleMoment}, \eqref{eqn:apndx_kneeMoment}, and \eqref{eqn:apndx_hipMoment} are consistent for both the stance leg and swing leg. For the swing leg, the ground reaction forces and moments are zero, i.e., $\boldsymbol{F}_G = \boldsymbol{0}$ and $\boldsymbol{M}_G = \boldsymbol{0}$. Translation accelerations and angular accelerations of the leg segments are extracted from the optical marker measurements. The corresponding anthropometry information of the participant such as segment length, mass, and location of COM can be found in Chapter 4 of~\cite{winter2009biomechanics}. We used estimate ratios for the joint moment calculation in this study.

\section{Data availability}
\label{data}
Supplementary dataset to this article can be found online at \url{http://dx.doi.org/10.17632/jgdpjrf584.2}.

\bibliographystyle{asmems4}

\bibliography{Ref_Chunchu_20223JanRev}

\end{document}